# Inference Networks and the Evaluation of Evidence: Alternative Analyses

**David A. Schum**
School of Information Technology and Engineering
School of Law
George Mason University
Fairfax, VA. 22032-4444
dschum@osf1.gmu.edu

## Abstract

Inference networks have a variety of important uses and are constructed by persons having quite different standpoints. Discussed in this paper are three different but complementary methods for generating and analyzing probabilistic inference networks. The first method, though over eighty years old, is very useful for knowledge representation in the task of constructing probabilistic arguments. It is also useful as a heuristic device in generating new forms of evidence. The other two methods are formally equivalent ways for combining probabilities in the analysis of inference networks. The use of these three methods is illustrated in an analysis of a mass of evidence in a celebrated American law case.

## 1.0 DIFFERENT STANDPOINTS, DIFFERENT ANALYTIC METHODS

There have been rapid advances made in our computer-based technologies for performing complex probabilistic reasoning tasks. Thanks to the work of many persons, including ones I will mention enroute, we are now able to draw conclusions based on complex probabilistic arguments in the form of *inference networks*. However, study of such networks is certainly not a recent venture. I will describe some work done as far back as 1913 on the construction and analysis of inference networks; but until the early 1970s no one paid much attention to this work. Like many other tasks, the construction and analysis of probabilistic inference networks can be performed in different ways and to suit different purposes. Choice of an analytic method depends on our perspectives, frames of reference, or *standpoints*.

For several years, my esteemed colleagues in field of law, Professors Terence Anderson and William Twining, have been arguing how advisable it is to clarify the standpoint according to which an analysis has been performed. As they mention in their work: *The Analysis of Evidence* [1991], the standpoint a person adopts strongly influences the ideas and evidence one will generate during the process of discovery, the arguments one will then construct based on this evidence, and the inferential force one will assign to these arguments. In declaring a standpoint, a person describes the role(s) he/she is playing, the objectives underlying the analysis, the process being analyzed, and the stage in this process at which the analysis is being performed. Anderson and Twining further argue that failure to declare these standpoint elements, up front, causes many difficulties for the person presenting an analysis as well as for members of an audience trying to assess its merits. In some cases the method a person employs and describes may seem unusual until this person announces the standpoint that led to the employment of this method. Following are some elements of standpoints that led to the methods I will decsribe and that led Jay Kadane [Carnegie-Mellon University] and I to employ them in a case study involving a large mass of evidence in a celebrated American law case. Two of the methods we used might seem quite unconventional, but they may also be necessary, or at least desirable, in many situations.

### 1.1 SOME OBJECTIVES IN STUDIES OF EVIDENCE AND INFERENCE

The major item on my research agenda for the past 35

years has been study of the properties, uses, and discovery of evidence in probabilistic reasoning. I have recently given an account of some of what I have learned in these studies [Schum, 1994]. Those of us working on early inference network studies in the 1960s had very little to draw upon in our studies. We were able to gain only a few relevant insights from the literature in probability, statistics, philosophy, and logic. One day it occurred to me that recorded experience and scholarship in the field of law might supply insights about evidence which we then lacked. I began to read works concerning evidence law and quickly realized what an important legacy this body of literature offers for anyone interested in probabilistic reasoning in any context, not just in law. Among other reasons, the adversarial nature of trials in our Anglo-American system of laws has forced scholars and practitioners in law to give very careful attention to the major credentials of evidence: its *relevance*, *credibility*, and *inferential [or probative] force*.

I paid particular attention to the work of an American evidence scholar named John Henry Wigmore [1863-1943]. Wigmore is certainly the most prolific and arguably the most profound scholar of evidence in the history of American jurisprudence. In the field of law, Wigmore is most noted for his 12-volume work on evidence [cited in my references as *Wigmore on Evidence*]. In this massive work Wigmore attends primarily to the admissibility of evidence. Not so widely known among jurists, but of greatest importance in the studies I will describe, is his inference-related work: *The Science of Judicial Proof: As Given by Logic, Psychology, and General Experience, and Illustrated in Judicial Trials* [1937]. While reading *Science of Judicial Proof*, I was astonished to observe how many answers Wigmore provided to questions about evidence and inference that were troubling us at the time.

From Wigmore's work I was first alerted to the many subtleties or complexities that can reside just below the surface of inferences we might believe to be quite simple. He verified my suspicion that, except in contrived situations, all probabilistic inference involves chains of reasoning often containing many links. In the 1960s we were using the term *cascaded inference* to describe such situations. Wigmore used the term *catenated inference*. From Wigmore's work I began to see how evidence items, alone and in combination, could be usefully categorized without regard to their substance or content and also to observe the different roles evidence plays in probabilistic reasoning. Finally, I was particulary enthusiastic about his analytic and synthetic methods for making sense out of *masses of evidence*.

These methods involve constructing arguments in defense of the relevance and credibility of different discernible forms and combinations of evidence. I believe Wigmore to have been the very first person to study the process of constructing and analyzing what today we call inference networks. He began these studies 86 years ago [Wigmore, 1913]. In Section 2.1 I describe a small portion of one of his inference networks.

Wigmore's works suggest other objectives such as probabilistic studies of various recurrent forms and combinations of evidence, which I have described elsewhere [Schum, 1994, 66-130]. Briefly, the forms identify whether the evidence is tangible, testimonial in various ways, is missing, or refers to accepted facts from authoritative records. Various recurrent combinations of evidence refer to different patterns of evidential harmony and dissonance as well as to different patterns of evidential synergism and redundancy. Each form and combination of evidence has a unique structural identity.

In these studies I made use of likelihood ratios from Bayes's rule to study how the inferential force of various forms and combinations of evidence depends on the likelihood ingredients of these likelihood ratios I derived to correspond to arguments based on various forms and combinations of evidence. In the process, I observed how the concept of *conditional nonindependence* is the major vehicle in conventional probability for capturing evidential subtleties or complexities. In Section 3.2 I provide some examples of the likelihood ratios I have studied for various evidential situations. My essential research strategy was to perform sensitivity analyses on the likelihood ratios I identified. It was from these sensitivity analyses that I learned additional things about the many subtleties that are associated with evidence and that are so commonly overlooked. Von Winterfeldt and Edwards once referred to this work as Bayes's rule meeting "ungodly inferences" [von Winterfeldt & Edwards, 1986, 163-204].

In many probabilistic studies of evidence it is very useful to have likelihood ratio equations because they help to explain more easily how the inferential force of evidence varies in response to changes in necessary ingredients of these equations. Given the nonlinearity of these equations and their many ingredient likelihoods, there are always surprises. In some cases, what seem to be major changes in ingredient likelihoods produce very small changes in a likelihood ratio. In other cases, however, exquisitely small changes in likelihood ingredients can produce drastic changes in a likelihood ratio. As the forms and combinations of evidence I examined became more complex, so did the likelihood ratios I had to derive in order to capture this complexity. Eventually, the task of deriving these likelihood ratios became unmanageable. I was greatly assisted by the doctoral dissertation work of Anne Martin who, to my knowledge, was the first person to program a computer so that it can behave as if it knows what likelihood ratio is appropriate to the structures associated with possible forms and combinations of evidence. Her program is called CASPRO [Martin, 1980].

For all his other accomplishments, Wigmore was no probabilist. Though he appreciated the fact that the linkages among elements of his inference networks were probabilistic in nature, he was only able to express these linkages in words, or in terms of *fuzzy probabilities*, as we would say today. Further, he provided no insights about how all the verbally-stated probabilities on his networks might be combined. Lotfi Zadeh was not around in Wigmore's time to show him how this might be accomplished. The emergence of computers naturally stimulated interest in developing efficient computational methods for aggregating or propagating probabilities in inference networks. The works of Pearl [1982, 1988], Lauritzen and Spiegelhalter [1988] and others hastened the development of commercially available software systems for the probabilistic analysis of inference or belief networks. I became interested in using these systems in probabilistic studies of inference networks constructed according to Wigmore's analytic and synthetic methods; an example follows in Section 3.3.

## 2.0 THREE METHODS OF ANALYSIS

Here are three methods of analysis that I will describe, the first of which is entirely structural in nature.

### 2.1 WIGMORE'S ANALYTIC AND SYNTHETIC METHODS

Suppose we have some existing mass of evidence and wish to construct arguments from this evidence to hypotheses being entertained. By such arguments we hope to establish the relevance, credibility, and inferential force credentials of items in this mass. Such situations occur very frequently in law, history, intelligence analysis, and in other contexts. Wigmore pondered the problem of establishing these credentials for masses of evidence items having a variety of different properties. He understood that arguments we construct from evidence to hypotheses or matters to be proven are all exercises in imaginative reasoning; different persons might construct quite different arguments from the same evidence. Constructing an argument from evidence to some hypothesis is essentially to provide a plausible and defensible chain of reasoning from the evidence to this hypothesis. Each link in such a chain identifies a possible source of uncertainty that may lurk between the evidence and what we are trying to prove from it. Wigmore argued that it would be very wise indeed for an attorney to construct arguments in advance of a trial or other settlement so that these uncertainties could be recognized and any disconnects [non sequiturs] in this argument could be identified and remedied.

Wigmore's work, although many decades old, is certainly not obsolete. His insights about probabilistic arguments based on evidence deserve serious attention. In fact, more recent works on the construction and analysis of arguments say many of the same things Wigmore did, but without any awareness of his earlier work [Toulmin, 1958; Toulmin, Reike, & Janik, 1984]. Inferences in legal affairs, whether civil or criminal, have an interesting structural characteristic that is illustrated below in Figure 1. In referring to hypotheses, Wigmore used the term *probandum* [matter to be proven]. In any civil or criminal case the *ultimate probandum* [$\Pi_u$] is the major hypothesis, civil claim, or criminal change at issue in the case at hand. For example, $\Pi_u$ might be: "Defendant X committed first degree murder in the slaying of Y on 30 January, 1999". Our substantive laws state exactly what elements or points must be proven [at some forensic standard such as "beyond reasonable doubt"] in order to sustain this ultimate probandum. These points or elements, shown in Figure 1 as $\Pi_1$, $\Pi_2$, $\Pi_3$, and $\Pi_4$, Wigmore called *penultimate probanda*. Different civil complaints or criminal charges of course have different penultimate probanda. In a charge of first degree murder, for example, the penultimate probanda are: $\Pi_1$ = Victim Y was killed; $\Pi_2$ = Defendant X killed Y; $\Pi_3$ = X intended to kill Y; and $\Pi_4$ = X fashioned this intent beforehand [i.e. the killing was premeditated].

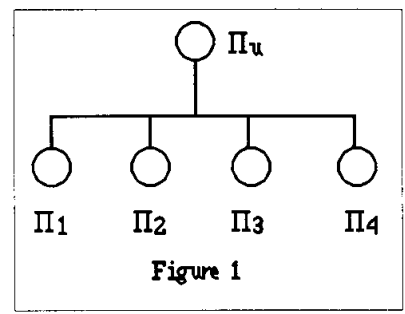

Figure 1

What all of this says is that inferences in legal contexts are well-structured at the top. Our substantive laws specify exactly what points must be proven in order to sustain some ultimate probandum. This is an inferential luxury that is not provided in any other context I know about. There are no "substantive laws" known to me in fields such as science, medicine, history, or intelligence analysis that tell us what penultimate matters we must prove in order to prove some major or ultimate hypothesis.

It is these penultimate probanda that supply touchstones for establishing the *relevance* of evidence. Wigmore understood that evidence can be relevant on some penultimate probandum in two ways. First, evidence is said to be *directly relevant* on a penultimate probandum if a plausible chain of reasoning can be constructed that links the evidence to the penultimate probandum. But other evidence can be *indirectly relevant*, or is *ancillary* evidence, if it bears on the strength or weakness of links [arcs] in chains of reasoning set up by directly relevant evidence. In short, ancillary evidence is evidence about other evidence and its strength or weakness. We might also term such evidence *meta-evidence* since it is evidence about evidence. I will have more to say about ancillary evidence as I proceed.

The interesting and difficult task, of course, is to construct arguments that link items in some mass of evidence to penultimate probanda such as those shown in Figure 1. Here is a task that rests on our imaginative reasoning coupled with critical thinking. Our chains of reasoning must be free of non sequiturs or disconnects. Further, as Wigmore noted, we must not leave any evidence out of our analysis that could be shown to be directly or indirectly relevant. Either disconnects in our arguments or evidence left unaccounted for invite the criticism of opponents. Wigmore understood how difficult it would be for most people to construct probabilistic arguments from a mass of evidence. To assist in this task, he devised an analytic and synthetic method for constructing inference networks that link our evidence to major hypotheses.

The analytic part consists of recording, on what Wigmore called a *key list*, propositions that define what we would today term the *nodes* on an inference network. Such nodes define the evidence, major hypotheses, and interim hypotheses appearing in arguments or chains of reasoning. In addition, Wigmore also suggested that we record inductive generalizations that provide a warrant or a license for reasoning from one node to another. One role of ancillary evidence is to test whether these generalizations in fact apply in the particular inference at hand. The synthetic part consists of drawing a chart showing the inferential relationships among the nodes identified on the key list. A small portion of one of Wigmore's inference networks is shown in Figure 2 below.

In Figure 2, evidence items are indicated by the ∞ symbol. Interim links in chains of reasoning, such as 7, are indicated by circles. The symbols on the arcs [such as < ] indicate the verbally-expressed strength of the probabilistic linkage between one node and another. The numbers identify propositions given on Wigmore's key list. The major hypothesis in this example concerns whether a victim, one Moses Young, died of poison he was allegedly given by a defendant named Oliver Hatchett.

Notice that the arcs on this diagram go <u>upward</u> from evidence to hypotheses. Wigmore emphasized that the reasoning of interest is inductive in nature, from evidence to hypotheses. I will have more to say about this matter, since some of my colleagues have often chided me for having my arcs going in the "wrong " direction on inference networks I have constructed.

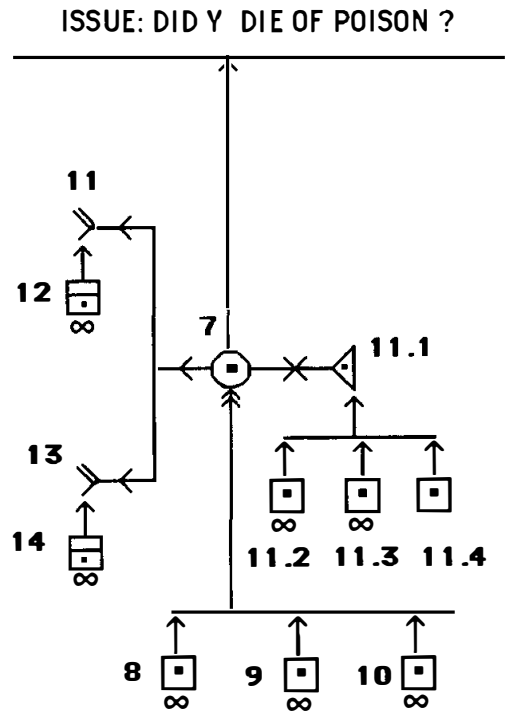

Figure 2

## 2.2 PROBABILISTIC ANALYSES

Most but not all of my studies of evidence have involved a Bayesian interpretation of what is meant by the inferential *force*, *weight*, or *strength* of evidence. I have also been interested in studying interpretations of these concepts in other systems of probability such as L. J. Cohen's Baconian probabilities and the Dempster-Shafer system of belief functions [e.g. Schum, 1986; 1991; 1994, 200-269]. The example of likelihood ratio studies of the force of evidence that I will describe in 3.2 captures an important evidential subtlety or complexity associated with evidential *redundance*.

In complex analyses, inference networks can be constructed according to a variety of different standpoints. In many situations we form an inference network as a model of how some complex process may work or unfold. Using imagination and background knowledge, we generate a model for this process that has many variables [nodes] indicating important sources of uncertainty. We also generate patterns of probabilistic influence by arcs indicating the direction of these influences among the nodes. As we know, all inference networks are *directed acyclic graphs* [incidently, Wigmore knew nothing about DAGs, but all of his networks have their properties].

In some but not all works on Bayesian inference networks the arcs are said to represent avenues of *causal* influence among nodes [Pearl, 1988, 117]. As I have noted elsewhere [Schum, 1994, 173-179], the concepts of *relevance* and *causality* are not the same; Bayes's rule works equally well in situations in which a relevance but no necessary causal influence can be discerned among nodes on a belief network. Inference networks, as models of complex processes, arise for a variety of different reasons. In some military applications, for example, an inference network can provide a basis for predicting future events. Other inference networks have been constructed for diagnostic purposes or simply to better understand some complex phenomenon. Inference networks constructed for these purposes may have any number of root nodes and so are unlike the ones I have constructed that typically involve just a single root node. Several different software systems are now available to perform the difficult task of aggregating probabilities in complex inference networks. In work I now describe, we made use of ERGO™ [Noetic Systems Inc.]

## 3.0 A CASE STUDY: THE SACCO AND VANZETTI EVIDENCE

Jay Kadane and I employed the three methods just described in an analysis of an existing mass of evidence. To illustrate these three methods we chose the trial and post-trial evidence in a law case decided many years ago but which still excites great controversy [Kadane & Schum, 1996]. On July 13, 1921 Nicola Sacco and Bartolomeo Vanzetti were convicted of first degree murder in the robbing and shooting of a payroll guard named Allesandro Berardelli in South Braintree, Massachusetts on April 15, 1920. After an unsuccessful six-year appeals process, they were executed by electrocution on August 23, 1927. There is little doubt that Sacco and Vanzetti were implacable anarchists [e.g. Avrich, 1991]. What is still in doubt is whether they were guilty of the crime with which they were charged. This case has been rightly called: *The case that will not die* [Ehrmann, 1969]. As recently as 1983 ballistics tests were still being performed on the firearms evidence in this case. The case of Sacco and Vanzetti is arguably the ranking *cause celebre* in all of American legal history. On some accounts the conduct of their trial represents the most serious miscarriage of justice in American legal history. Even if their trial was fair, there are many lingering doubts about the guilt of Sacco and Vanzetti.

Kadane and I considered both the trial evidence [consisting of over 2200 pages of transcript] and evidence made available in many works published in the years after their trial. We considered a total of 395 items of evidence. Of this total, 226 items came from the trial and 169 items were generated after the trial. Here are the three penultimate probanda in this case, each of which the prosecution was obliged to prove to the forensic standard: *beyond reasonable doub*t:

$\Pi_1$: Berardelli was killed,
$\Pi_2$: Berardelli was killed during a robbery of a payroll he was carrying,
$\Pi_3$: It was Sacco, with the assistance of Vanzetti, who intentionally shot Berardelli during a robbery of the payroll.

What we have in this case is a species of homicide called *felony murder*. If Sacco and Vanzetti had just robbed Berardelli at gunpoint and had not killed him, they would, if convicted, have been sentenced to life imprisonment. Because they were also alleged to have killed Berardelli, they faced the death penalty. In short, the prosecution was never required to prove premeditation. The evidence presented by the prosecution on $\Pi_1$ and $\Pi_2$ was not contested by the defense; the only contested issue in this case was $\Pi_3$.

The three penultimate probanda just shown represent the "relevance hooks" on which we sought to place the 395 items of evidence we considered. Of this total, we believed 164 items to be directly relevant on one or the other of the penultimate probanda in this case and 231 items we believed to be indirectly relevant or ancillary in nature. The prosecution had three major lines of argument on $\Pi_3$, the only contested issue in this case.

One line of argument was based on identification evidence and concerned what Sacco and Vanzetti were allegedly doing before, during, and after the crime. The second line of argument was based on firearms evidence and concerned the weapons Sacco and Vanzetti were carrying with them when they were arrested, the bullets taken from Berardelli's body, and various other items. The final line of argument concerned what is known in law *as consciousness of guilt* [*mens rea*]. It was alleged by the prosecution that the behavior of Sacco and Vanzetti when they were arrested, and at trial, demonstrated that they knew they were guilty of the crime with which they were charged.

### 3.1 WIGMOREAN ANALYSIS: FINDING AN INFERENTIAL HOME FOR 395 ITEMS OF EVIDENCE.

We first employed Wigmore's analytic and synthetic methods in constructing arguments from the evidence on $\Pi_1$, $\Pi_2$, and $\Pi_3$, described above. To do so required us to adopt the standpoint of the prosecutors in this case. Neither the prosecution nor the defense were ever required to generate detailed arguments in defense of the evidence they sought to introduce. The collection of arguments we generated from the 164 items of directly relevant evidence constituted our inference network. Having generated this network of arguments, we were able to find links in these chains of reasoning to which each one of the 231 remaining items of evidence were indirectly relevant or ancillary. Our entire belief network consists of 28 sectors, each of which concerns a specific substantive issue or line of argument on one or the other of the three penultimate probanda in this case. Figures 3 and 4 below concern the directly relevant and ancillary evidence in one such sector, that I will describe in a moment. Wigmore's methods are uniquely suited for the analysis of an existing mass of evidence. But they are also very helpful in discovery-related activities when new possibilities and new evidence are to be generated.

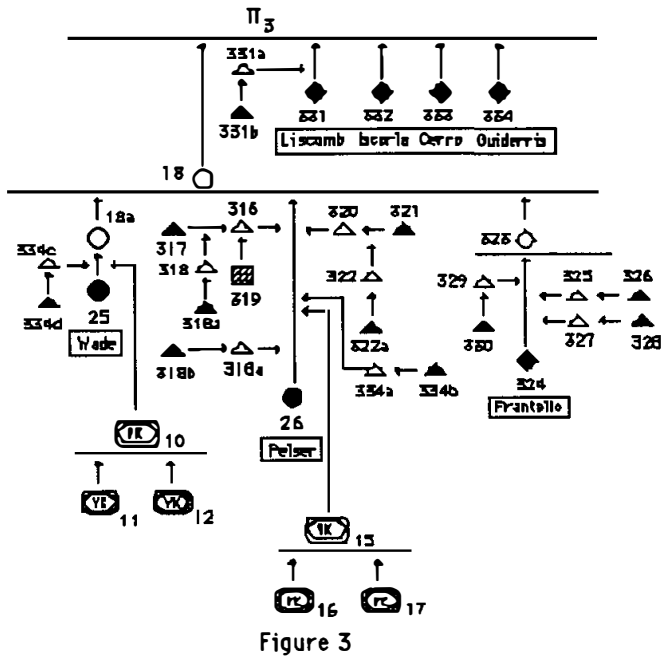

Figure 3

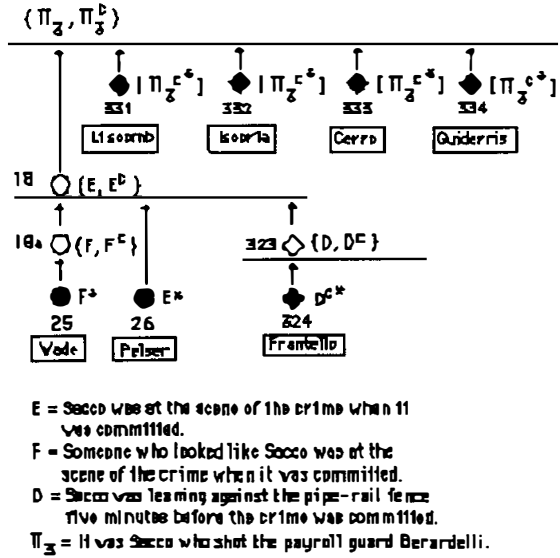

Figure 4

Figures 3 and 4 above show a sector of our Wigmorean inference network that concerns what is called *concomitant* identification evidence regarding Sacco. This evidence concerns what Sacco was allegedly doing or not doing at the time of the crime. There are seven items of evidence, directly relevant on $\Pi_3$, in this sector.

Items 25 and 26 concern the testimony of two persons, Pelser and Wade, who were believed to be the prosecution's "star" witnesses. Items 324, 331, 332, 333, and 334 come from witnesses Frantello, Liscomb, Iscorla, Cerro, and Guiderris who were thought to be the defense's "star" witnesses. All the rest of the items shown in Figure 3 as either black triangles or the one cross-hatched box are items of ancillary evidence bearing on the credibility of the witnesses who gave directly relevant testimony. Figure 4, which I will explain later, is the same as Figure 3 with the ancillary evidence omitted. Here comes an important point about our probabilistic analyses of the evidence in this case.

There are no statistics available to support probability judgments on the arcs for the directly relevant evidence in our analysis. The events of concern in this case, as well as those in most other law cases, are *singular, unique*, or *non-replicable*. The events about which we have testimonial evidence from these seven witnesses either occurred or did not occur at one time in the past. We cannot play the world over again n times in order to see how many times these events happened. If this is so, then what supports any probability *judgments* at the arcs from directly relevant evidence? The answer is that the ancillary evidence we have is an appropriate basis for judging likelihoods on these arcs. For example, what we learn about Pelser and what he was doing at the time of the crime forms the ancillary basis for credibility-related assessments such as his hit and false-positive probabilities. Notice that chains of reasoning from the ancillary evidence point to but do not connect to arcs on chains of reasoning from the directly relevant evidence. As I noted earlier, ancillary evidence is evidence about other [directly relevant] evidence and its inferential strength or weakness.

Some might believe that we should connect all evidence on the same network and ignore the distinction between directly and indirectly relevant evidence. I believe this would be quite a mistake. In the first place, it would require a prodigious imagination to construct arguments from some of the ancillary evidence in this case to the penultimate probandum at issue, $\Pi_3$. Second, this being a situation in which there are no statistics to support probability judgments, if we take away the ancillary evidence we have no basis for assessing the likelihoods Bayes's rule requires us to assess. My charting of ancillary evidence corresponds with Ron Howard's identification of *evocative information* that he says supports probability and value judgments on influence diagrams [Howard, 1989, 907]. In other works I have discussed at some length the treatment of ancillary evidence on belief networks [Schum, 1994, 187-192; Kadane & Schum, 1996, 85-88].

### 3.2 LIKELIHOOD RATIOS AND THE "STAR" WITNESSES

I have another reason for showing you the inference network sector in Figure 3: Kadane and I used it to illustrate the likelihood ratio method of analysis that I described in Section 2.2. Before I tell you about the likelihood ratios we derived for the directly relevant evidence in Figure 4, I need to tell you more about what the seven witnesses named above testified. For the prosecution, Pelser, who worked on the second floor of the Rice & Hutchins shoe factory in front of which the crime occurred, said he heard shots fired in the street below and saw, from a window, a man he identified at trial as Sacco at the crime scene. Wade, who was on the street about 150 feet from where the crime occurred, would only say that he saw a man who "looked like" Sacco at the scene when the crime occurred.

Four defense witnesses, Liscomb, Iscorla, Cerro, and Guiderris, were just across the street from the crime scene; all testified that it was not Sacco who shot the payroll guard. Several witnesses testified that, shortly before the crime, there were two men leaning against a pipe-rail fence in front of the Rice & Hutchins shoe factory and that these two men attacked Berardelli and another payroll guard as they passed. Defense witness Frantello testified that he saw these two men five minutes before the crime and that Sacco was not one of them.

In Figure 3, item 317 refers to the testimony of another defense witness named Constantino who worked with Pelser at the Rice & Hutchins shoe factory. Constantino testified that, when the shots were fired, several workers, including Pelser, dove under a workbench. Here is a good example of the ancillary evidence I have mentioned. What Pelser was doing at the time of the crime, by itself, has no conceivable direct relevance on $\Pi_3$, that Sacco shot the payroll guard. However, this testimony is indirectly relevant since it bears on Pelser's credibility. Pelser cannot have seen Sacco at the scene of the crime when it happened if he was under a workbench at the time. Other ancillary evidence in Figure 3 was further damaging to Pelser's crediblity.

Kadane and I derived likelihood ratios for the evidence given by each of the seven witnesses whose directly relevant evidence is shown in Figure 4. The reader interested in examining all of these likelihood ratios can consult our work [Kadane & Schum 1966, Table 6.3, p 199]. At present, I only have enough space to tell you about the likelihood ratios for Pelser's and Wade's testimonies that are shown below. They provide an example of the subtleties that can be captured in this form of Bayesian analysis.

Note: The events in these expressions are defined in Figure 4.

For Pelser's Testimony E*:

$$L_{E^*} = \frac{P(E|\Pi_3)[h_p - f_p] + f_p}{P(E|\Pi_3^c)[h_p - f_p] + f_p}, \text{ where } h_p = P(E^*|E); f_p = P(E^*|E^c), \text{ for Pelser.}$$

For Wade's Testimony F*, Given Pelser's Testimony E*:

$$L_{F^*|E^*} = \frac{P(E|E^* \Pi_3)[P(F|E) - P(F|E^c)](h_w - f_w) + P(F|E^c)(h_w - f_w) + f_w}{P(E|E^* \Pi_3^c)[P(F|E) - P(F|E^c)](h_w - f_w) + P(F|E^c)(h_w - f_w) + f_w}, \text{ where } h_w = P(F^*|F); \text{ and}$$

$f_w = P(F^*|F^c)$ for Wade; and where:

$$P(E|E^* \Pi_3) = \frac{P(E|\Pi_3)h_p}{P(E|\Pi_3)h_p + P(E^c|\Pi_3)f_p}, \text{ and } P(E|E^* \Pi_3^c) = \frac{P(E|\Pi_3^c)h_p}{P(E|\Pi_3^c)h_p + P(E^c|\Pi_3^c)f_p}.$$

In $L_{E^*}$ for Pelser, $P(E|\Pi_3) = 1.0$ since, if Sacco did kill Berardelli, then he must have been at the scene of the crime. What is to be judged is $P(E|\Pi_3^c)$: How likely is Sacco to have been at the scene of the crime if he did not kill Berardelli? The values $h_p$ and $f_p$ are credibility-related hit and false-positive probabilities for Pelser. Now, the likelihood ratio for Wade's testimony, $L_{F^*|E^*}$, is quite interesting since the force of Wade's testimony depends, in part, on Pelser's credibility. The subtlety or complexity here concerns possible *evidential redundance*. If we believed Pelser to be perfectly credible, and thus also believed Sacco to have been at the scene of the crime, then Wade's testimony can have no inferential value. If Sacco was at the scene of the crime, then someone who looked like Sacco must have been there. Wade's testimony has value to the extent that Pelser is not credible. The terms $P(E|E^*\Pi_3)$ and $P(E|E^*\Pi_3^c)$ tell us how much inferential value associated with Sacco being at the crime scene is "left over" after we have Pelser's testimony. The term $P(F|E) = 1.0$ since, if Sacco was at the scene, then someone looking like him must also have been at the scene [there was never any suspicion that the persons involved in this crime wore disguises]. What requires judgment is $P(F|E^c)$; How probable is it that someone who looked like Sacco was at the crime scene, given that Sacco himself was not there? The values $h_w$ and $f_w$ are credibility-related hit and false-positive probabilities for Wade.

We employed likelihood ratios such as these to tell a variety of different stories about the inferential force of the evidence in the case of Sacco and Vanzetti. The stories we told, based on sensitivity analyses, involved different probabilistic beliefs that might be held about the ingredients in these likelihood ratios. These likelihood ratios supply endings to the many different stories we told about the evidence in this case. However, some of the stories we wished to tell about the force of this evidence were far too complex for likelihood ratio analysis. To find endings for these stories we used the ERGO™ Bayesian inference network software.

### 3.3 ERGO AND BULLET III

Shown in Figure 5 is an inference network, analyzed using ERGO™, for an issue that arose concerning the firearms evidence used by the prosecution against Sacco. A forensic surgeon named Dr. Magrath, who performed the autopsy on Berardelli, testified that he extracted four bullets from Berardelli's body and identified each one with a Roman numeral he scratched on the base of them. He also testified that the bullet he had marked III caused Berardelli's death. A 32-caliber bullet marked with a III on its base was shown at trial and identified as Exhibit 18. When he was arrested, Sacco was carrying a 32-caliber Colt automatic, which was shown at trial as Exhibit 28. The prosecution offered evidence that Exhibit 18 [Bullet III] was fired through Sacco's Colt [Exhibit 28]. Both the prosecution and defense were allowed to test-fire other 32-caliber bullets through the Colt automatic identified as Sacco's. The prosecution claimed that the markings on the test-fired bullets matched those on Exhibit 18. Event M at

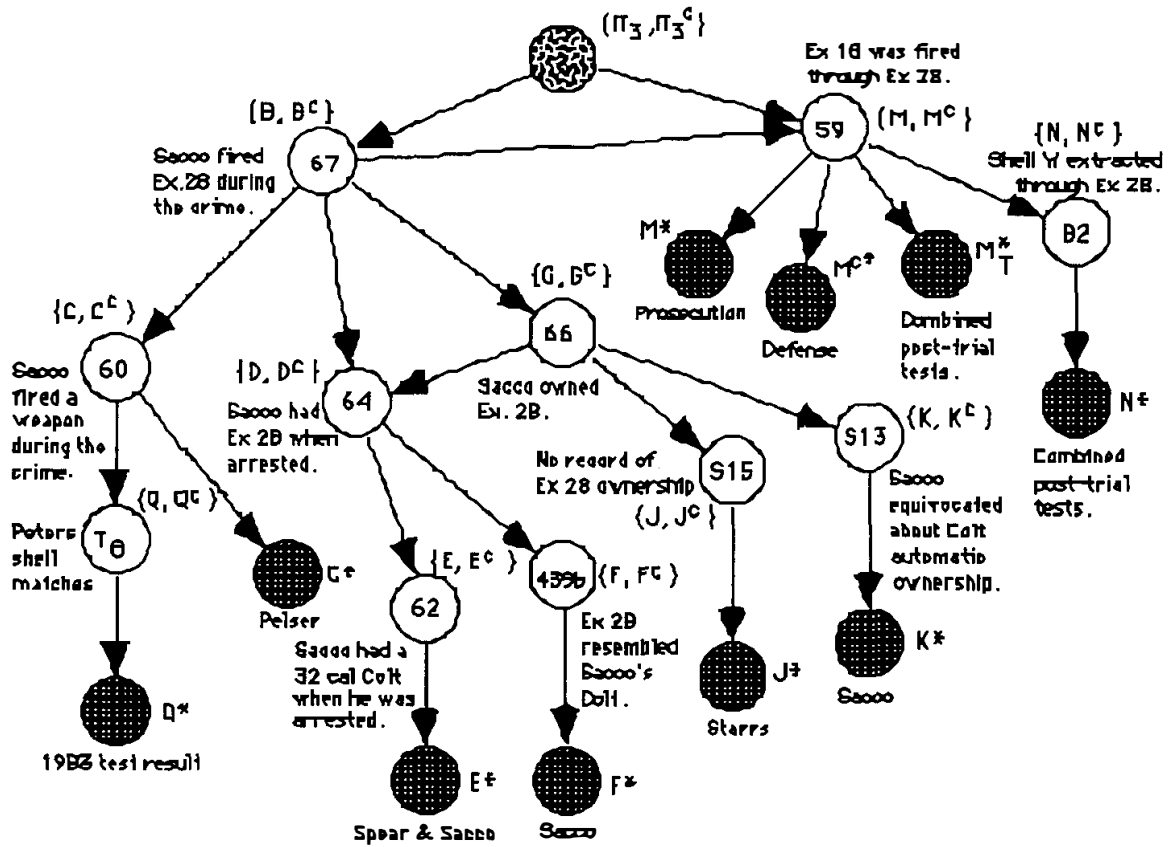

Figure 5

Node 59 in Figure 5 says that Exhibit 18 was fired through Exhibit 28. If true, event M would favor $\Pi_3$, that Sacco did shoot Berardelli. However, by itself, this event would not be very persuasive since someone else might have fired Sacco's Colt during the crime. The "star" witness Pelser appears again with his further testimony that Sacco fired *a weapon* during the crime [event C at Node 60]. Other evidence concerned Sacco's possible ownership of Exhibit 28 [event G at Node 66], and whether Exhibit 28 was the same Colt taken from Sacco when he was arrested [event D at Node 64]. Events C, D, and G, if true, converge to suggest event B at Node 67, that Sacco fired Exhibit 28 during the crime. Here we have an example of an important inferential subtlety capturable in Bayesian terms. Events B and M are *synergistic* in their influence on $\Pi_3$. Together, these events have more inferential force favoring $\Pi_3$ when they are taken together than they do if considered separately or independently. In short, events B and M seem to be nonindependent, conditional on the truth of $\Pi_3$. There is another way to express this synergism. Event M, that Exhibit 18 was fired through Exhibit 28, has more force on $\Pi_3$ if we also consider event B, that Sacco fired this Exhibit 28 during the crime.

## 4.0 IN CONCLUSION

Bayesian Inference networks, now being employed with greater frequency, can be grounded on evidence having different forms and appearing in different combinations. In this paper I have suggested that different methods for the analysis of these networks are useful depending upon the standpoint of the person(s) performing the analysis. Although Wigmore's original methods are cumbersome, and now seem quaint, in more "user-friendly" forms they provide many insights about complex probabilistic reasoning and enforce a certain discipline in the construction and analysis of inference networks. In applications of Bayesian inference networks, when there is no statistical backing for the likelihoods these methods require us to assess,

584    Schum

Wigmore's methods show us a reasonable basis for subjective assessments of these likelihoods. This is one important role that ancillary evidence or evocative information [to use Ron Howard's term] plays.

In reasonably simple situations in which it is necessary to explain why or how some result was obtained, the likelihood ratio method I described is very useful. Having expanded likelihood ratios at hand that show how their ingredients should be combined facilitates explanations and also reveals the many subtleties or complexities may lie just below the surface. One trouble with existing Bayesian belief network software systems is that their aggregation or propogation algorithms are buried below the surface. But these systems are certainly helpful and indeed are necessary in analyses whose complexity outruns our inclinations or abilities to write likelihood ratio expressions for the force of evidence on which a belief network is grounded. In some situations, such as in the case study I have described, we can profitably employ all of these methods of analysis.